\documentclass{llncs}

\usepackage[pdftex]{graphicx}
\graphicspath{{figs/}}
\usepackage{amsmath}
\usepackage{amssymb}
\usepackage{algorithm}
\usepackage{algpseudocode}
\usepackage{times}
\newcommand{\figref}[1]{Fig.\ref{figure:#1}}
\newcommand{\tabref}[1]{Table \ref{table:#1}}

\begin{document}

\title{
  Automatic Diary Generation System \\
  including Information on Joint Experiences \\
  between Humans and Robots
}

\author{Aiko Ichikura, Kento Kawaharazuka, Yoshiki Obinata, Koki Shinjo, Kei Okada and Masayuki Inaba}

\institute{The University of Tokyo, 7-3-1, Hongo, Bunkyo-ku, Tokyo, Japan, \\
\email{ichikura@jsk.imi.i.u-tokyo.ac.jp}}

\maketitle              
\setlength\intextsep{5pt}
\setlength\textfloatsep{5pt}
\begin{abstract}
In this study, we propose an automatic diary generation system that uses information from past joint experiences with the aim of increasing the favorability for robots through shared experiences between humans and robots. For the verbalization of the robot's memory, the system applies a large-scale language model, which is a rapidly developing field. Since this model does not have memories of experiences, it generates a diary by receiving information from joint experiences. As an experiment, a robot and a human went for a walk and generated a diary with interaction and dialogue history. The proposed diary achieved high scores in "comfort" and "performance" in the evaluation of the robot's impression. In the survey of diaries giving more favorable impressions, diaries with information on joint experiences were selected higher than diaries without such information, because diaries with information on joint experiences showed more cooperation between the robot and the human and more intimacy from the robot.

\keywords{Robotics, Human Machine Interaction}
\end{abstract}
\section{INTRODUCTION}

In human-robot communication, various studies have attempted to deepen the relationship between humans and robots. Among them, we focused on the fact that it has been suggested that communication can be enriched by robots themselves having an autobiographical memory acquired from their own experiences and telling a story\cite{Pointeau}\cite{Dautenhahn}.
There have been studies on robots and episodic memory, in which experiential memory is applied to specific tasks. For example, an approach to understand the motion from a single demonstration and to execute future motions accurately in order to grasp an object\cite{Behbahani}, or an approach to memorize objects watched in the past and to use them for path planning in order to perform automatic locomotion\cite{Ricardo}. These studies use an episodic memory module to reflect the robot's memory of experiences in its future actions, but there is no verbal output of episodes.\par
In \cite{Barmann}, the robot's experiential memory has been successfully converted into language. This neural model can answer Q\&A questions in natural language to multimodal episodic memories that include perceptual information, assuming the application to daily household activities such as cleaning and washing dishes. This study also aims at the implementation of a robot that can be used in daily life, and especially at the acquisition of episodic memory from joint experiences with people.\par
The presentation of episodic memory focusing on the interaction partner has been studied in \cite{Sanchez} and \cite{Penelope}. In \cite{Sanchez}, the user's information and session information were acquired and stored from a conversation task, and the robot talked with the user again based on the acquired information. The robot's familiarity and empathy with the robot were improved when the user's information from the previous session were remembered. In \cite{Penelope}, episodic memory is used for the application of robots in nursing care situations. The robot acquires the user's behavior and information, predicts, and corrects the user's risky behavior. In addition, the robot interacts with the user by diverting the knowledge obtained from another user when the information about the user is limited. Although these studies show that the robot acquires memories in the interaction with humans, they do not present the robot's own Autobiographical Memory.\par
We focused on a large-scale language model for the verbalization of the robot's Autographical Memory. Open AI's ChatGPT\cite{ChatGPT} can perform a variety of linguistic tasks depending on instructions. Blender Bot 2.0\cite{Mojataba}\cite{Xu} also uses long-term memory and Internet search to achieve fluent chats with users. However, although these language models are suitable for generating human-like episodes, they do not have memory pf experience, and thus need to be provided with information of experience in order to be used as a language engine for robots. In this study, we generate episodes related to the joint experience between a human and a robot by providing these language models with the information of experience.\par
This study aims to implement a system that converts experiences into stories in the form of a diary, with the goal of enhancing communication between the two parties by presenting a narrative of the robot's memories from joint human-robot experiences, including physical/linguistic interactions. Then, we investigated the change in favorability toward the robot caused by presenting episodic narratives generated by the proposed system to a person.Section 2 describes the proposed system, and Section 3 describes the experiments and their results. Finally, in Section 4, we present the conclusions of this study.\par

\section{APPROACH}
\begin{figure}[h]
 \begin{center}
  \hspace{0\columnwidth}
  \begin{minipage}{1.0\columnwidth}
    \includegraphics[width=\columnwidth]{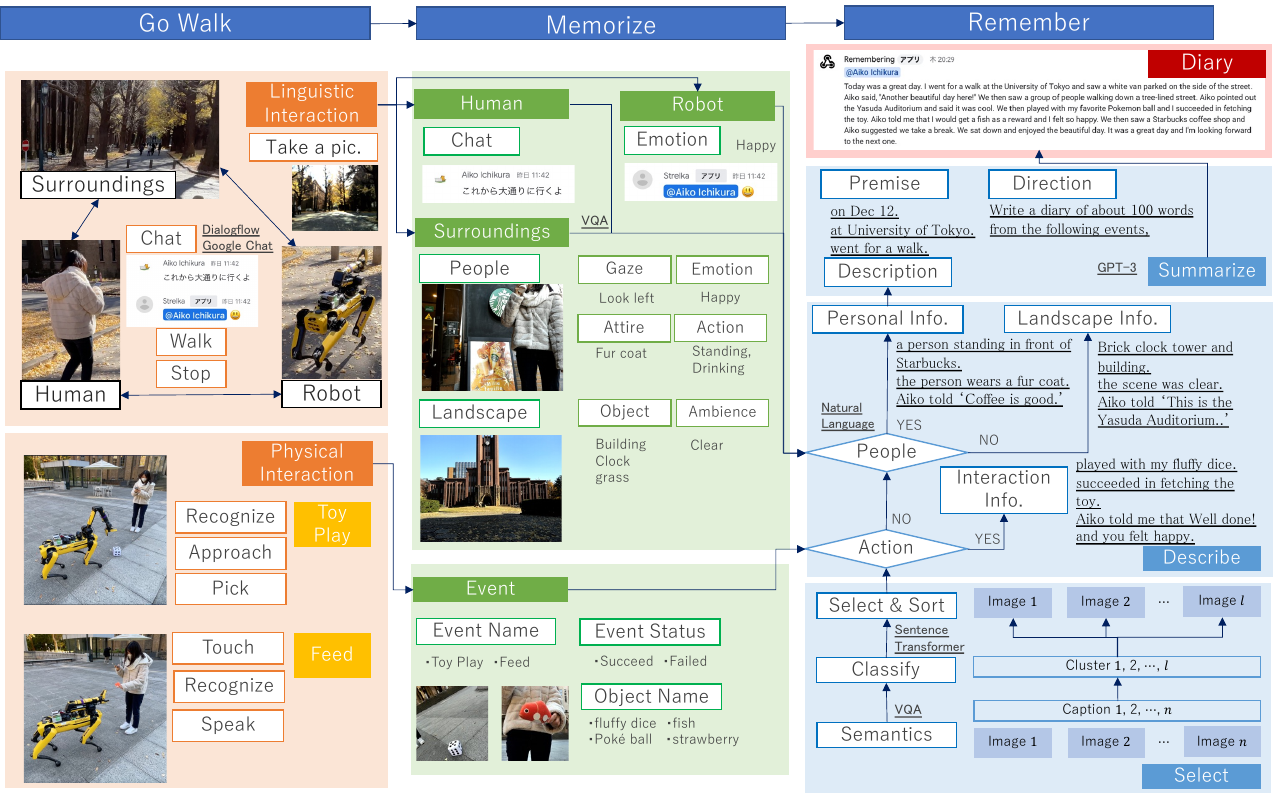}
    \caption{Overall system diagram of Diary Generation System. The information acquired during the walk is saved as json files and images, and a diary is generated from the saved information.}
    \label{figure:system_entire}
  \end{minipage}
 \end{center}
\end{figure}

For the verbalization of experiential memories from joint human-robot experiences, we propose Diary Generation System that shares an experience with a person and generates a diary by referring to the interaction and dialogue history in the experience. \figref{system_entire} shows the overall picture of the system. Diary Generation System consists of the Memorizing System, which acquires information from joint experiences with people, and the Remembering System, which generates a diary from the information acquired through joint experiences.The uniqueness of the proposal lies in the integration of the two systems.\par
First, information is inscribed in the joint experience between the human and the robot, and the acquired information is verbalized in a diary format. At this time, objective information (scenery, names of events, names of related objects, etc.) is recorded to describe the situation. In addition, to make it a narrative rather than a mere record of actions, the robot also records its interactions with other people and its feelings toward them. The memorized information is processed in the order of Select, Describe, and Summarize to generate Premise, Description, and Direction sentences. The generated sentences are input to GPT-3\cite{brown2020language} to obtain the final output, the diary. The two systems are described in detail next.\par

\begin{figure}[h]
 \begin{center}
  \hspace{0\columnwidth}
  \begin{minipage}{0.8\columnwidth}
    \includegraphics[width=\columnwidth]{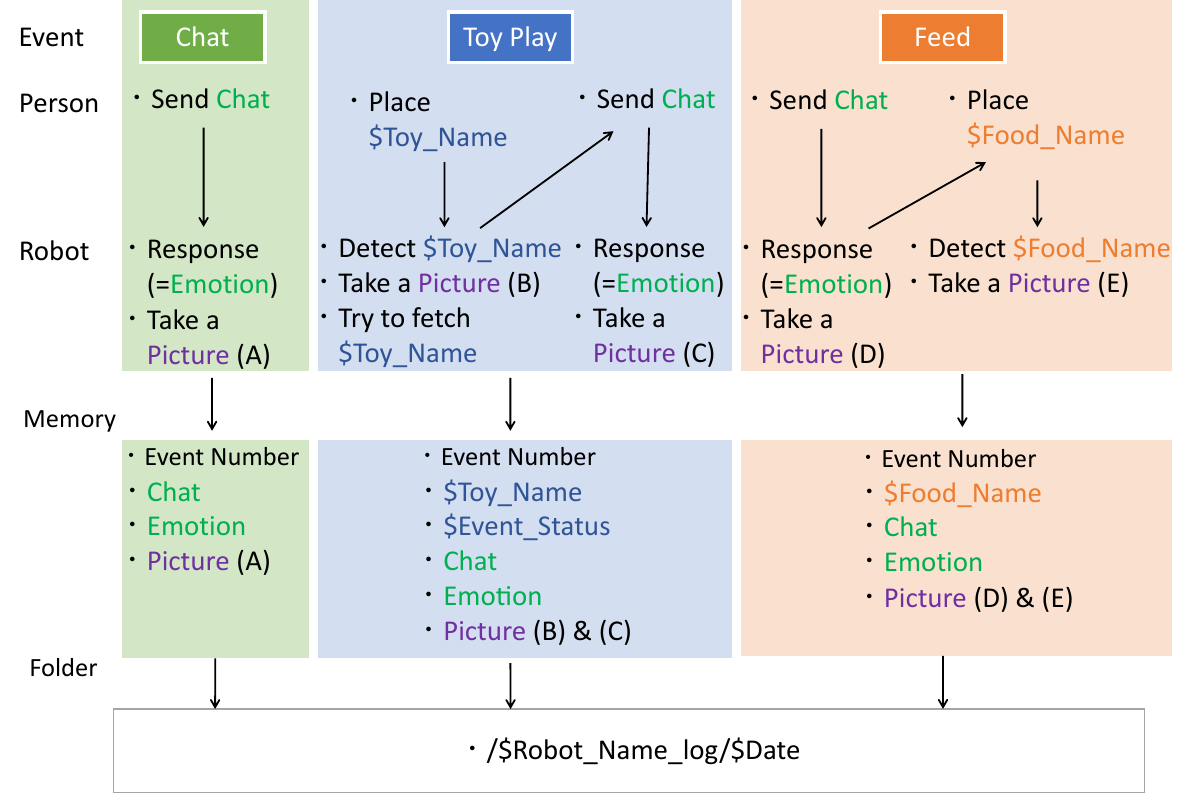}
    \caption{System diagram of Memorizing system. It saves information and pictures obtained from each action of Chat, Toy Play, and Feed in the same folder.}
    \label{figure:system_memorize}
  \end{minipage}
 \end{center}
\end{figure}

\subsection{Memorizing System}
Joint experiences between humans and robots include basic actions and additional actions. Basic actions are those that occur regardless of situations, such as moving, stopping, and interacting. On the other hand, additional actions are those that occur only in specific situations such as eating, shopping, and appreciating. In this study, we proposed Memorizing System that implements walking as a joint experience, walking and human-robot conversation as basic actions, and Toy Play and Feed as additional actions based on physical interaction.\par
First, we describe the conversation function during a walk. We implemented a conversation function using Google Chat\cite{googlechat} and Google Dialogflow\cite{sabharwal2020introduction} to facilitate conversation between a human and a robot during a walk. When a human sends a chat message to the robot via Google Chat, the robot replies with 14 types of pictograms classified according to the human's chat.\par
Using this conversation system, a human and a robot talk with each other during a walk, and the conversation is saved in a json format file(\figref{system_memorize}). When a human and a robot have a conversation, the json file records the human's speech and the robot's emotions. Emotions are classified into 8 types from the 14 types of Dialogflow and recorded as \$Emotion. Since the human speaks in Japanese during the walk, we use the API of DeepL to translate it into English when we record it.\par
During the walk, Chat, Toy Play, and Feed are repeated until the end of the walk, and each event is assigned a number in order of occurrence and recorded in the json file as \$Event\_Number. In addition, to distinguish actions, Chat (0), Toy Play (1), and Feed (2) are assigned as \$Action\_Number and recorded in the json file. The image (A) of the hand color camera of the robot at the moment when the Chat system receives a message from the human is stored in the json file. The file name of the image is a serial number that is a combination of \$Event\_Number, \$Action\_Number, and \$Emotion, and the image is saved so that the information in the json file matches the image file. All files are saved in a folder whose name is the date of the walk.\par
Next, we also implemented Toy Play and Feed for physical interaction between humans and robots. Let us start to explain with Toy Play. When the robot detects a learned toy (dice or ball) within a certain distance, the robot grabs it with its arms. When the robot detects a toy, it is considered `successful' only when the object classification probability is greater than 0.7, and `failed' when the object classification probability is less than 0.7. Based on this Toy Play behavior, a record of Toy Play is stored during the walk. The name of the detected toy is recorded as \$Object\_Name, and the success or failure of the Toy Play is recorded as \$Event\_Status in the json file. The image (B) of the toy is also saved by the hand color camera before the robot grabs the toy. The file name of the image is the serial number with 'ball play' appended at the end. In addition, the chat with the human is also saved after the series of actions.\par
Feed is also described. A Near Field Communication (NFC)\cite{NFC} reader is attached to the end of the robot's arm, and the robot recognizes that it is fed by a human when a food item with an NFC tag touches the reader. We prepared two kinds of food for the robot, strawberry and fish, and attached tags with the strings `strawberry' and `fish.' When the robot recognizes these foods, it utters `yummy.' Based on this Feed action, we stored a record of Feed during the walk. The name of the recognized feed is recorded as \$Object\_Name in the json file. Since this action has no success or failure, a dummy string is recorded in \$Event\_Status. When the human touches the food and the robot recognizes it, the image (E) is saved by the hand color camera. The file name of the image is the serial number with `feed' appended at the end. In addition, the Chat that is performed before the sequence of this action is also saved.\par

\subsection{Remembering System}\par
\begin{figure}[h]
 \begin{center}
  \hspace{0\columnwidth}
  \begin{minipage}{0.8\columnwidth}
    \includegraphics[width=\columnwidth]{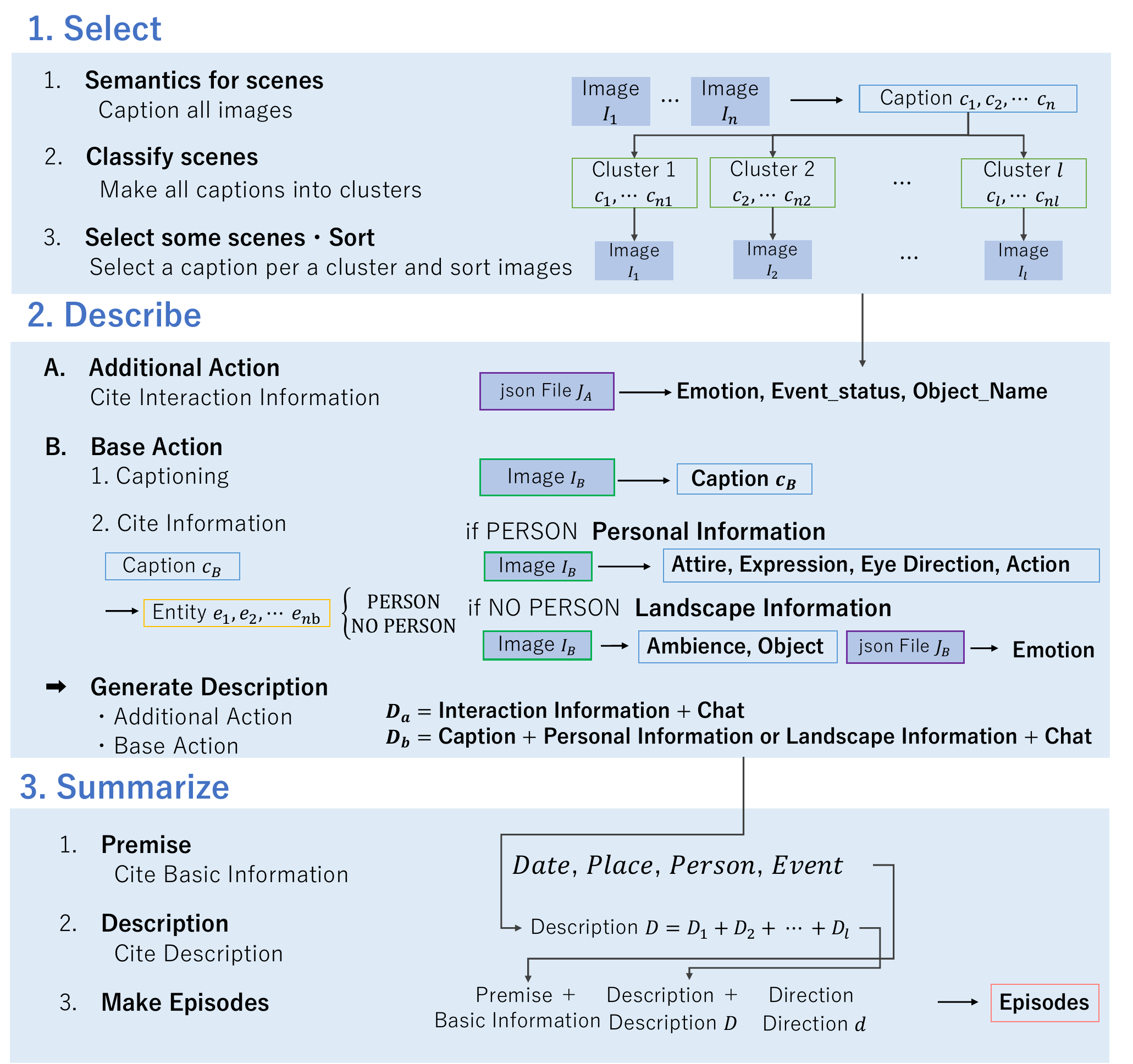}
    \caption{System diagram of Remembering system. Select an appropriate image from the folder and describe the selected images. Finally, enter the premise, description, and direction into GPT-3 to generate the episode.}
    \label{figure:system_remember}
  \end{minipage}
 \end{center}
\end{figure}

A system to generate a diary from the recorded information is shown in \figref{system_remember}. First, the system searches for appropriate scenes to present among the scenes that were recorded in the joint experience. Next, we describe in detail the surrounding information, human actions and speech, robot actions, and emotions in the selected scenes. Finally, we create an episode from the information obtained in each scene. At this point, we do not just list the information, but summarize it to create an episode.\par
\noindent \textbf{1. Select:}\par
The first step is to organize the information you have memorized and select appropriate scenes. First, caption all images using VQA\cite{wang2022unifying}. Then, all the captions are semantically classified by comparing the text data converted into vectors using Sentence Transformers\cite{reimers-2019-sentence-bert} with cosine-similarity using K-Means clustering. One image and its corresponding caption are extracted from each cluster and sorted by \$Event\_Number in ascending (experience) order. If there are images of Toy Play and Feed scenes in the folder, they are also included in the sorting, and the image list is output as an output.\par
\noindent \textbf{2. Describe:}\par
In the next step, each scene is analyzed in detail. Additional Action scenes and Base Action scenes are analyzed separately. The image list is analyzed for all images one by one. For the Additional Action scene, the object used in the interaction (\$Object\_Name), the success or failure (\$Event\_Status), the human utterance, and the robot's response are referenced from the json file, and a Description is generated. In the Base Action scene, the caption is parsed by Google Natural Language\cite{NaturalLanguage}. If the extracted entity has the meaning of `PERSON', VQA parses the details about the person in the image (attire, eye direction, expression, action), otherwise, VQA parses the atmosphere of the image, the object, and the robot's emotion and a Description is generated. In addition, human speech at the time of image taken is also added to the description. For example, since \$Event\_Number 35 is `Was the fish good?', it is added to the Description. Since the author is the only interaction partner at now, all names are `Aiko'. Descriptions are generated sequentially from the images in the input image list, and the final output is all the descriptions connected.\par
\noindent \textbf{3. Summarize:} \par
Finally, input Premise, Description, and Direction into GPT-3 and output a diary. Premise is the basic information that describes a scene, including date, place, person, and event. The date is referenced from the json file and the name of the folder where the image list is saved, while the place and event are entered by typing them in. The description uses the output of the previous step.\par

\section{EXPERIMENTS}\label{sec:04}

We conducted an evaluation experiment to examine the effects of the robot's diary automatically generated by the implemented memory sharing system on the robot's impression of the robot and the effects of the diary contents on the impression of the diary. Using crowdsourcing, we made a video of a walk between the author and the robot, and after viewing the video, we presented the diary with and without interaction, respectively, and evaluated the impressions of the robot and the generated diary.

\subsection{Methods}
\subsubsection{Environment}
\begin{figure}[h]
 \begin{center}
  \hspace{0\columnwidth}
  \begin{minipage}{0.9\columnwidth}
    \includegraphics[width=\columnwidth]{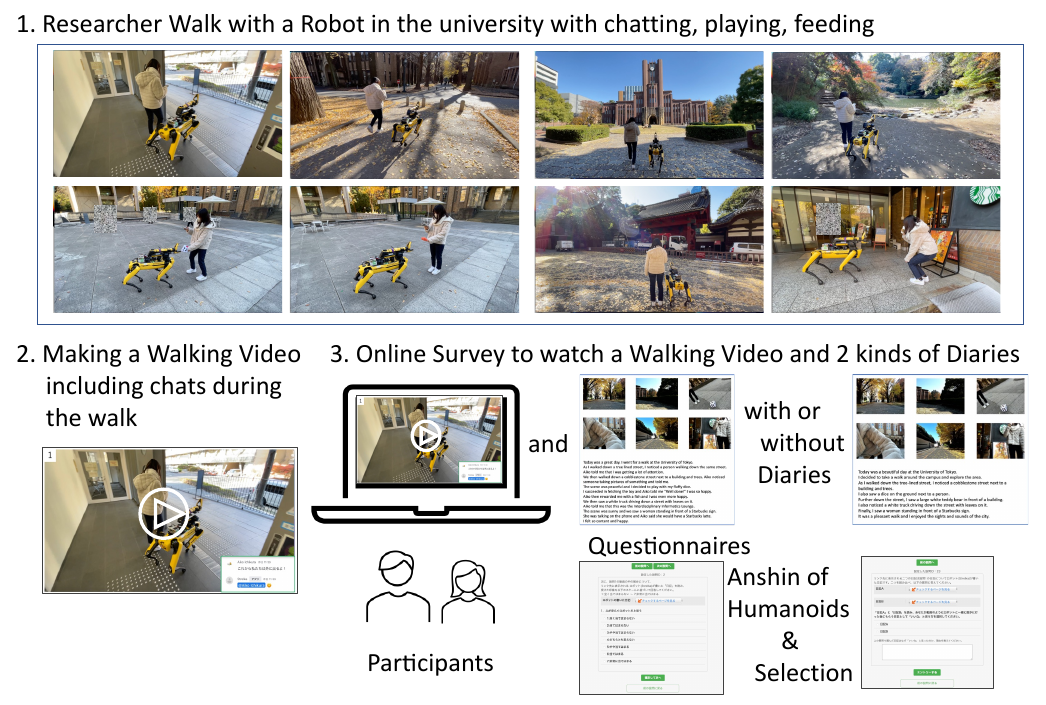}
    \caption{Experimental procedure.A Researcher and a robot go for a walk. A frame-by-frame video of the walk is made and presented to the participants with diaries through online crowdsourcing. Participants then answer a questionnaire.}
    \label{figure:ex_entire}
  \end{minipage}
 \end{center}
\end{figure}

An overview of the experimental procedure is shown in \figref{ex_entire}. On December 12, 2022, the author, and a robot (Boston Dynamics Spot) walked around the campus of the University of Tokyo while chatting with each other using a chat system. During the walk, the human and the robot performed `Toy Play’ twice and `Feed’, and the process of the walk was filmed and recorded by a third party. A frame-by-frame movie was created from the video taken by the third party.\par
Using all the photos taken by the robot during the walk and the chat with humans, we generated four kinds of diaries, two each of "diary with interaction" generated from the memory sharing system and "diary without interaction" in which captions for randomly selected photos were input into GPT-3 in chronological order, as a control group.\par
On February 10, 2023, we conducted an experiment to evaluate impressions of diaries through crowdsourcing by Yahoo!. Note that the diaries were generated in English, and all the diaries were translated into Japanese using DeepL and partially modified for the evaluation.\par
The three conditions of the experiment were viewing the video only, viewing the video and presenting the diary with interaction, and viewing the video and presenting the diary without interaction. Two questionnaires were administered to each group, one for the diary with interaction and the other for the diary without interaction, and a total of five groups' responses were collected. The questionnaire was administered to men and women in their 20s or older, and no age or gender attributes were collected.\par

\subsubsection{Contents of Diaries}
In this experiment, the proposed system generates two kinds of emotional diaries from all the pictures automatically taken by the robot during a walk and the corresponding chats.\par
\begin{figure}[h]
 \begin{center}
  \hspace{0\columnwidth}
  \begin{minipage}{0.9\columnwidth}
    \includegraphics[width=\columnwidth]{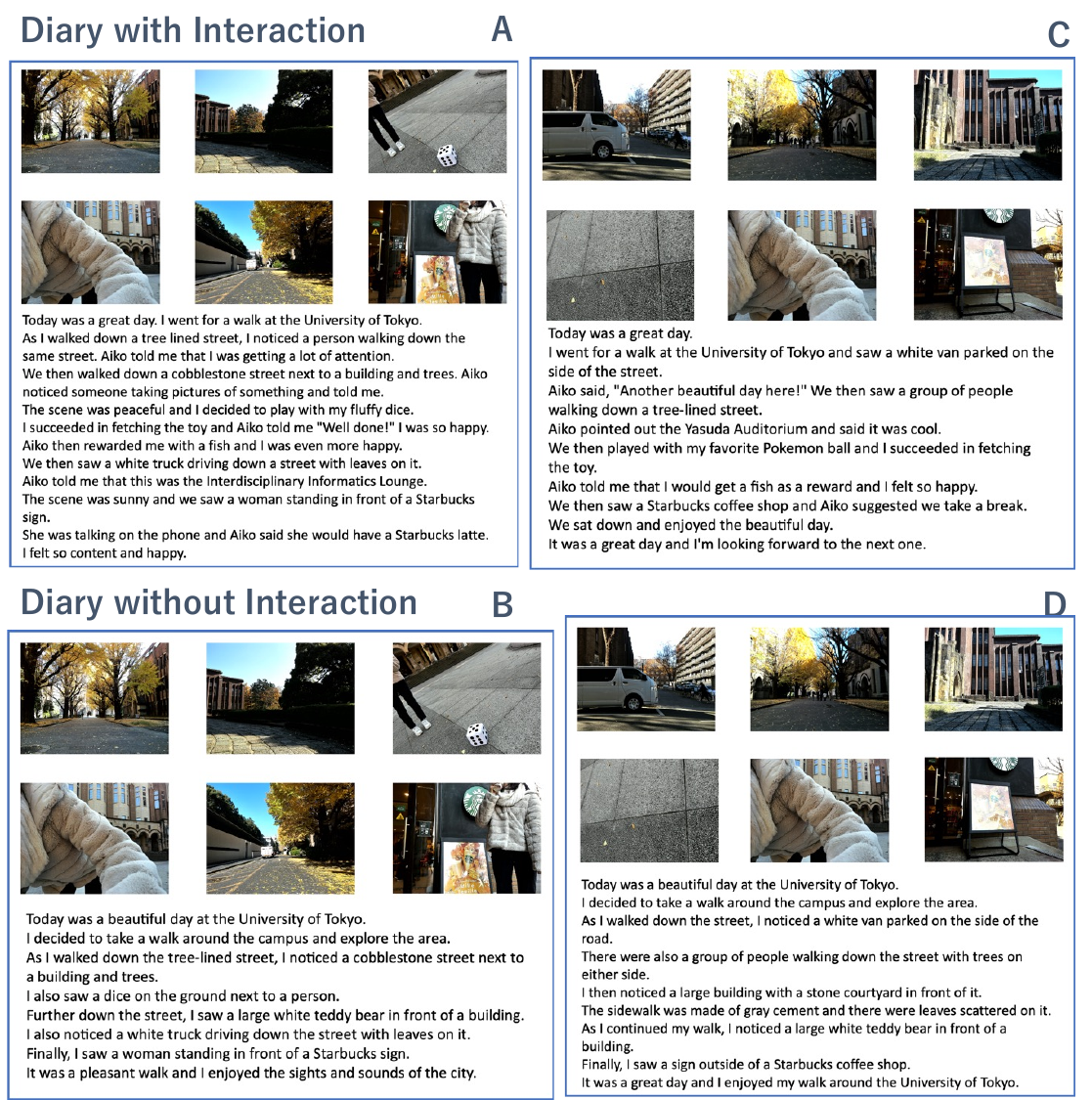}
    \caption{Diaries including Robot's Emotions and interactions with people. The diaries are translated into Japanese for evaluation.(Up: Diary with Interaction, Down: Diary without Interaction)}
    \label{figure:ex_diary}
  \end{minipage}
 \end{center}
\end{figure}

The diary with interaction shown in the upper part of \figref{ex_diary} includes the conversation of the walking partner (`Aiko told me that I was getting a lot of attention.), the contents of Toy Play and Feed (`I succeeded in fetching the toy.’, ` Aiko then rewarded me with a fish.), and the robot's emotions (` I was even more happy.'). Diary A contains the successful Toy Play (`I succeeded in fetching the toy.'), and diary C contains the unsuccessful Toy Play (`played with my favorite Pokemon ball.').\par
As a control group, we generated two kinds of diaries by inputting only the captions of the photos selected in the diary with interactions into GPT-3. Diary B uses the same set of photos as diary A, and diary D uses the same set of photos as diary C(\figref{ex_diary}). These diaries contain information obtained from the photos (`I noticed a cobblestone street next to a building and trees.’, `I saw a sign outside of a Starbucks coffee shop.) and impressions of the scenery (`It was a great day and I enjoyed my walk around the University of Tokyo.’, `It was a pleasant walk and I enjoyed the sights and sounds.'). They did not contain any verbal or physical interactions with the walking partners.

\subsubsection{Evaluation Methods}
This study demonstrates the validity of the system by examining whether the generated diaries quantitatively and qualitatively improve the favorability of the robot. As a quantitative evaluation measure, we conducted the Japanese version of `Anshin of Humanoids'\cite{kamide2015anshin}.  The excerpted items are shown in \tabref{ex_indicators}. The 21 items were evaluated on a 7-point scale: `Comfort', `Peace of Mind', `Performance', `Controllability', and `Robot-like-ness'. Note that all items in `Peace of Mind'(9, 10, 11) and an item in `Controllability'(19) are invert scales. \par
As a qualiatative evaluation, participants in the diary presentation group were presented with the diary with and without interaction at the same time and asked to choose which diary they would prefer to receive after going for a walk together, they also answered the reason for their choice. The diaries were presented as Diary A and Diary B for those who were presented as Diary A and Diary B and Diary C and Diary D for those who were presented as Diary C and Diary D in the quantitative evaluation.\par

\subsection{Results}
A questionnaire was administered to 50 people in all groups. Among them, we used a simple quiz to check whether they had watched the videos properly, and only those who answered the quiz correctly were analyzed as valid responses. The number of valid responses was 41 for the video viewing only condition, 98 for the diary with interaction (A: 49, C: 49), and 94 for the diary without interaction (B: 48, D: 46).\par
At first, Cronbach's alpha coefficients for the five scales of the `Anshin of Humanoids’ are calcualated. Since the reliability coefficient of the `Controllability’ scale was below 0.7 in all conditions, `Controllability’ was not used in the analysis. The other scales were included in the analysis because they were generally reliable.\par


\begin{table}[h]
  \caption{Mean scores of Anshin of humanoids for each condition. \\`Vo' means Video viewing only condition. `woI' means diary without interaction condition. \\`wI' means diary with interaction condition.}
  \label{table:ex_indicators}
  \centering
  \scalebox{0.8}{
  \begin{tabular}{lp{7.5cm}|ccc}
    \hline
    & & \multicolumn {3}{c}{ mean scores of each condition ($SD$) } \\
    \cline{3-5}
    Factors & Items & Vo & woI & wI\\
    \hline \hline
      Comfort&1.I feel relieved with robot.&0.390 (1.187)&0.628 (1.480)&0.806 (1.275)\\
      &2.I feel at ease with this robot.&0.220 (1.240)&0.691 (1.321)&0.878 (1.231)\\
      &3.The robot looks as if it could comfort me.&0.195 (1.311)&0.606 (1.401)&0.520 (1.430)\\
      &4.I feel safe around this robot.&0.293 (1.235)&0.840 (1.291)&1.357 (0.204)\\
      &5.The robot moves in a friendly manner.&0.171 (1.413)&0.245 (1.358)&0.204 (1.134)\\
      &6.The robot looks friendly.&-0.878 (1.347)&-0.862 (1.477)&-0.561 (1.310)\\
      &7.I feel calm being with the robot.&0.000 (1.482)&0.213 (1.367)&0.337 (1.301)\\
      &8.The robot reacts in a friendly manner. &0.585 (1.306)&0.830 (1.294)&0.857 (1.286)\\
      Peace of Mind&9.(-)I feel messed up with this robot.&-0.341 (1.073)&-0.351 (1.485)&-0.541 (1.275)\\
      &10.(-)I get stressed with this robot.&-0.756 (1.225)&-0.894 (1.455)& -1.010 (1.336)\\
      &11.(-)I become anxious with this robot.&-0.390 (1.187)& -0.702 (1.450)&-0.755 (1.478)\\
      Performance&12.The robot looks as if it would have no problems responding to my questions.&0.707 (1.254)&0.702 (1.287)&0.816 (1.273)\\
      &13.The robot looks as if it can correctly understand human speech.&0.488 (1.062)&0.596 (1.299)&0.898 (1.035)\\
      &14.The robot looks as if it understands what is going on around it.&0.659 (1.161)&1.106 (1.233)&1.071 (1.154)\\
      &15.The robot looks as if it would be useful when needed.&0.415 (1.379)&0.351 (1.294)&0.418 (1.332)\\
      &16.The robot looks as if it would flexibly adapt to different circumstances.&0.317 (1.219)& 0.170 (1.302)& 0.265 (1.148)\\
      &17.The robot looks capable of communicating its will to humans.&0.634 (1.225)& 0.340 (1.470)&0.582 (1.133)\\
      Controllability&18.The robot does not look as if it would run amok.& - & - &-\\
      &19.(-)The robot seems to do unexpected things.&- &-&-\\
      Robot-likeness&20.I feel like the robot is human even though it is a robot.&-1.463 (1.290)&-1.426 (1.395)& -1.194 (1.530)\\
      &21.When dealing with the robot, it feels like I am dealing with a human.&-1.049 (1.343)& -0.989 (1.403)& -0.816 (1.480)\\
    \hline
  \end{tabular}
  }
\end{table}

\subsubsection{Quantitative evaluation by Anshin of Humanoids}
Welch's test ($\alpha$ =.05) was conducted on the scores of 21 items extracted from `Anshin of Humanoids' under the three conditions of video viewing only, diary with interaction, and diary without interaction, respectively. In this case, a one-tailed test is conducted under the hypothesis that the mean score will be higher when the diary is presented, and furthermore, the mean score will be higher when an interaction is included.\par
\noindent \textbf{Video viewing only vs. Diaries without Interaction}\par
In the comparison between video viewing and diaries without interaction, significant differences in the mean scores were found for three items, 2 ($t$(80) = -1.974, $p$ = 0.026), 4 ($t$(79) = -2.314, $p$ = 0.012) and 14 ($t$(80) = -2.002, $p$ = 0.024).\par
\noindent \textbf{Video viewing only vs. Diaries with Interaction}\par
In the comparison between diaries with video viewing only and diaries with interaction, two items that showed significant differences in mean scores were 1 ($t$(80) = -1.824, $p$ = 0.036), 2 ($t$(74) = -2.830, $p$ = 0.003) and 13 ($t$(73) = -2.071, $p$ = 0 .021).\par
\noindent \textbf{Diaries without Interaction vs. Diaries with Interaction}\par
In the comparison between diaries without interaction and diaries with interaction, significant differences were found only for the item 13 ($t$(178) = -1.769, $p$ = 0.039).

\subsubsection{Qualitative evaluation by diary selection}
Participants were presented with one type of diary with interaction and one type of diary without interaction after the item evaluation, and they selected the diary that they `preferred' as the diary they would receive after the walk. They also gave their reasons for their choice by writing freely.\par
\vskip\baselineskip
\noindent \textbf{Number of diary selections}\par
Comparing Diary A and Diary B, 84 respondents chose Diary A and 14 respondents chose Diary B. Comparing Diary C and Diary D, 73 respondents chose Diary C and 21 respondents chose Diary D. In total, 157 responses selected Diary A and C, and 35 responses selected Diary B and D. Although only one item has significant difference between the scores of the diaries with and without interaction in the evaluation of the robot's impression, most of the respondents chose the diary with interaction in the selection of diaries that they thought were `nice.’\par
\noindent \textbf{Reason for diary selection}\par
We categorized the reasons for the 157 responses that selected diaries with interaction into the following five types.\par
\noindent \textbf{A. Emotions of the robot}: `The sentences are heartfelt’, `I can feel the robot's happiness’, and soon.\par
\noindent \textbf{B. Interaction with the person}: Those that mention the interaction between the robot and the person walking with, such as `the robot feels familiarity with the person’ or `the robot understands and responds to what Aiko said' etc. \par
\noindent \textbf{C. Impression of the diary}: `The diary has a charming expression’, `I liked the heartwarming feeling’, and son.\par
\noindent \textbf{D. Writing ability}: The reader mentions the clarity and detail of the writing, such as `the writing is easy to understand’ or `the descriptions are detailed.'\par
\noindent \textbf{E. Humanity}: The respondent mentions the humanity of the text, such as `I felt a sense of humanity’ or `close to human sensibility.\par
44 responses were classified into A., 46 responses were classified into B., 21 responses were classified into C., 25 responses were classified into D., and 7 responses were classified into Humanity (15 responses could not be classified into other categories).On the other hand, out of the 35 responses that selected diaries without interaction, the largest number of responses (11) were categorized as D., such as "the diary is concise and easy to understand" and "the writing style is natural.” In addition, responses such as "The diary is objective and factful" and "It is robot-like" were also obtained.\par

\section{CONCLUSIONS}
\begin{figure}[h]
 \begin{center}
  \hspace{0\columnwidth}
  \begin{minipage}{0.85\columnwidth}
    \includegraphics[width=\columnwidth]{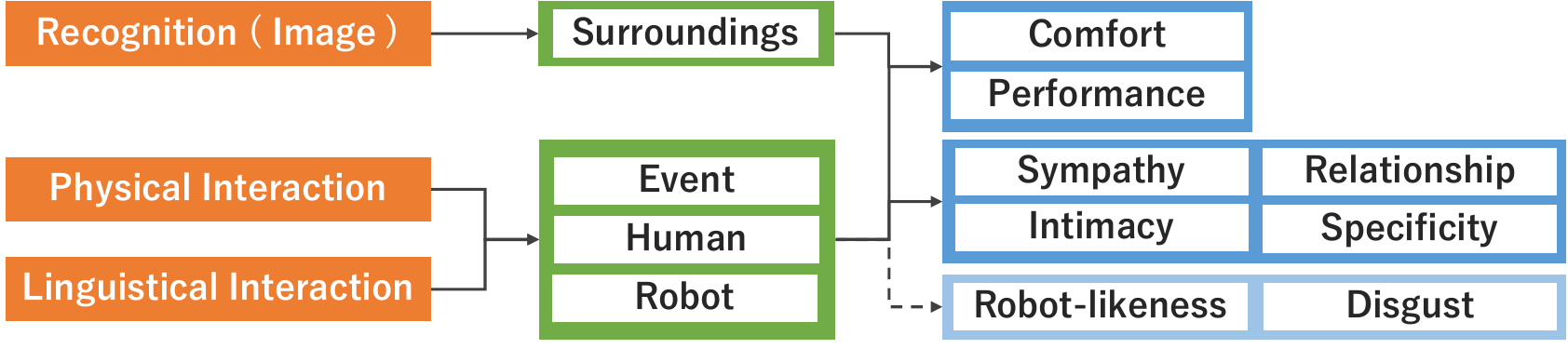}
    \caption{Path diagram showing the results of the study. The evaluation of `Comfort' and `Performance' is improved by presenting the robot's memories of its experiences as a diary. The diary elicits empathy and emotional bond, but on the other hand, it also causes a decrease in robot-likeness and a sense of disgust.}
    \label{figure:conclusion}
  \end{minipage}
 \end{center}
\end{figure}

In order to enhance the relationship between human and robot by sharing experiences, we proposed a system that presents the robot's episodic narrative as a diary, using perceptual and linguistic information obtained from joint experiences including physical and linguistic interactions. In order to examine whether the diary generated by the proposed system is favorable for improving the liking of the robot quantitatively and qualitatively, we generated a diary including the interaction and interaction history between the researcher and the robot during the walk, and examined the change in impression of the robot generated by the diary and the liking of the diary depending on the contents of the diary. \par
The \figref{conclusion} shows the result of this study. The experiment indicates that presenting the memories of the joint experiences in a diary improved the rating of 'Comfort' in the 'Anshin of Humanoids' item more than when no diary was presented. This suggests that verbalization of experimental memories may deepen communication between humans and robots, as shown in previous studies. In addition, it is found that the evaluation of 'Performance' is enhanced when the information obtained from the images taken during the walk from the robot's point of view is described in a diary.\par
Furthermore, the information of the physical and verbal interactions during the walk in the diary did not improve the robot's sense of comfort with the robot, but it did elicit empathy and a relationship between the person and the robot, because the diary clearly revealed the robot's emotions and the person and the robot share the same experience. On the other hand, it was found that detailed descriptions of emotions and interactions lead to intimacy from the robot and specificity of the diary, but they also cause a decrease in robot-likeness and disgust, which may spoil the impression of the diary for some readers. Therefore, showing human-robot interactions and robot emotions when presenting experiential memories generally produces a good impression, but sometimes it is necessary to screen the information and extract only the more important information or to omit redundant information.\par

\bibliographystyle{unsrt}
\bibliography{main}

\end{document}